# DeepCL: Deep Change Feature Learning on Remote Sensing Images in the Metric Space

Haonan Guo, *Student Member, IEEE,* Bo Du, *Senior Member, IEEE,* Chen Wu, *Member, IEEE,* Chengxi Han, *Student Member, IEEE,* Liangpei Zhang, *Fellow, IEEE.*

*Abstract*—**Change detection (CD) is an important yet challenging task in the Earth observation field for monitoring Earth surface dynamics. The advent of deep learning techniques has recently propelled automatic CD into a technological revolution. Nevertheless, deep learning-based CD methods are still plagued by two primary issues: 1) insufficient temporal relationship modeling and 2) pseudo-change misclassification. To address these issues, we complement the strong temporal modeling ability of metric learning with the prominent fitting ability of segmentation and propose a deep change feature learning (DeepCL) framework for robust and explainable CD. Firstly, we designed a hard sample-aware contrastive loss, which reweights the importance of hard and simple samples. This loss allows for explicit modeling of the temporal correlation between bi-temporal remote sensing images. Furthermore, the modeled temporal relations are utilized as knowledge prior to guide the segmentation process for detecting change regions. The DeepCL framework is thoroughly evaluated both theoretically and experimentally, demonstrating its superior feature discriminability, resilience against pseudo changes, and adaptability to a variety of CD algorithms. Extensive comparative experiments substantiate the quantitative and qualitative superiority of DeepCL over state-of-the-art CD approaches. The source code will be made available at https://github.com/HaonanGuo/DeepCL.**

## I. INTRODUCTION

High spatial-temporal resolution Earth observation techniques have facilitated the acquisition of multiple high spatial-resolution remote sensing (RS) images of the same area within a short period[1]. Leveraging multi-temporal Earth observations enables the monitoring of dynamics on the Earth's surface[2]. Change detection (CD) techniques aim to identify regions of change between images captured in the same geographical area but at different times. Automatic binary CD algorithms utilize bi-temporal image pairs as input and generate a change map by categorizing each pixel as either changed or non-changed[3]. CD has gained considerable attention in recent decades, due to its practical significance in various domains such as disaster response[4], urban planning[5], resource management[6], and agricultural monitoring[7].

The emergence of deep learning techniques has further propelled the development of automatic CD algorithms[8]. Early deep learning-based CD methods adopted a single-stream design, wherein bi-temporal images were simply concatenated or subtracted before being fed into the network, allowing seamless integration with existing semantic segmentation networks[9]. Despite simple modification of the network architecture, these methods yielded limited performance since shallow convolutional layers cannot extract informative features that conform to the distribution of individual raw images[3]. Moreover, these methods are susceptible to noise and variations in illumination, making them unlikely to outperform advanced CD algorithms[10]. Further studies adopted a dual-stream architecture for CD, employing either Siamese or pseudo-Siamese encoders to extract hierarchical representations from bi-temporal images independently. Temporal feature modeling was then performed to transform

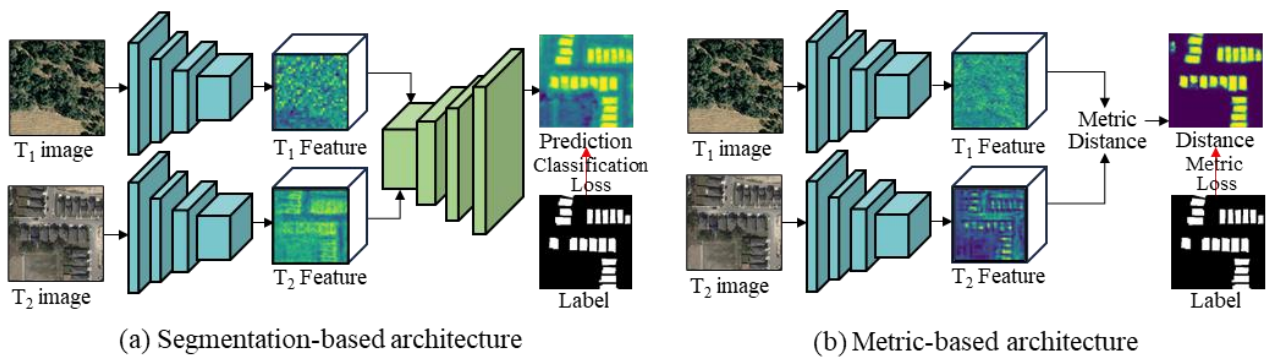

(a) Segmentation-based architecture                (b) Metric-based architecture

Fig.1 Architectures of the (a) segmentation-based change detection and the (b) metric-based change detection.

Haonan Guo, Chen Wu, Chengxi Han, and Liangpei Zhang are with the State Key Laboratory of Information Engineering in Surveying, Mapping and Remote Sensing, Wuhan University, Wuhan, China (e-mail: haonan.guo@whu.edu.cn, chen.wu@whu.edu.cn, chengxihan@whu.edu.cn, zlp62@whu.edu.cn ).

Bo Du is with the National Engineering Research Center for Multimedia Software, Institute of Artificial Intelligence, School of Computer Science and Hubei Key Laboratory of Multimedia and Network Communication Engineering, Wuhan University, Wuhan, China (e-mail: gunspace@163.com).

the bi-temporal features into a change map[11]. According to the principle of temporal modeling, dual-stream CD methods can be divided into two categories: (1) segmentation-based methods and (2) metric-based methods. In segmentation-based CD methods, bi-temporal features are fused and transformed into change features using convolution or absolute difference. Upsampling and convolution operations are employed to generate the change map from the fused features, as depicted in Fig. 1(a). The model parameters are optimized by minimizing the loss function between the predicted change map and the ground truth. A wide range of approaches have focused on enhancing the transformation of bi-temporal image features into change feature representations, such as incorporating attention mechanisms[12] and dense feature connections[13].

In contrast to segmentation-based methods that require additional model design for fusing bi-temporal features into change features, metric-based methods directly measure the distance between bi-temporal features in the metric space[14], as depicted in Fig.1 (b). Since features of the same object are expected to share similar representations in bi-temporal images, pixels with metric distances surpassing a predetermined threshold can be identified as changed regions. The parameters of metric-based models are optimized by minimizing the metric distance between bi-temporal features in non-changing regions and maximizing the metric distance in changed regions. During inference, the metric distance map is separated into changed and unchanged categories using a distance threshold. Further improvements in the metric-based methods have focused on refining metric losses[15] and feature aggregation[16].

Despite the affirming performance of DLCD achieved in varied CD-oriented applications, it still faces two limitations that need to be overcome. The existing segmentation-based CD methods suffer from insufficient temporal modeling due to the lack of effective feature optimization. In the commonly used Siamese network architecture, it is assumed that the same ground object at the same location will be represented similarly in the feature space. However, segmentation-based methods optimize their parameters implicitly using the loss calculated at the output space, lacking explicit constraints for Siamese network learning. Consequently, these methods have limited abilities to learn change representation due to inadequate semantic relations mining. On the other hand, metric learning-based methods directly optimize the Siamese network by utilizing the metric distance between bi-temporal image features. While these methods can explicitly optimize temporal semantic relations, ensuring that non-changing objects share similar feature embeddings and changed objects have different embeddings, they generate the final change map through a manually set threshold from the distance map. This approach is susceptible to pseudo-change detection caused by misregistration and illumination variation in remote sensing images. Considering the strengths and weaknesses of metric-based and segmentation-based DLCD methods, metric-based approaches effectively model the temporal correlation of features, which is lacking in segmentation-based methods.

Conversely, segmentation-based methods alleviate the issues of manual threshold selection and pseudo-change detection by directly learning the decision plane from annotations. It is therefore meaningful to explore a combination of the complementary advantages offered by metric-based and segmentation-based methods.

In this paper, we propose a Deep Change Feature Learning (DeepCL) framework for change detection using bi-temporal remote sensing images by integrating the strong temporal modeling ability of metric learning and the excellent fitting ability of segmentation methods. DeepCL incorporates a simple yet effective hard sample-aware contrastive loss ($L_{HSAC}$) to reweight the gradients of difficult samples in metric-space optimization, which remarkably exceeds the existing metric distance loss. With improved feature representation achieved by $L_{HSAC}$, a metric-induced decoder network is proposed to fuse bi-temporal features and generate the change map. We believe that our work provides the first step toward the interconnection of optimization in both metric space and classification space for improved CD performance. The main contribution of this paper includes:

1.  A Deep Change feature Learning (DeepCL) architecture is proposed to combine the strong temporal modeling ability of metric space optimization and the fitting ability of output space optimization. By enhancing the bi-temporal image representations through metric space learning, the resulting CD results generated by the decoder network exhibit superior precision and interpretability compared to existing CD methods.

2.  To enhance the temporal modeling ability in the metric space, a novel hard sample-aware contrastive loss is designed. The proposed metric loss function facilitates the model's focus on challenging samples and is proven more effective theoretically and experimentally with merely one line of code in implantation. Furthermore, it mitigates the subjectivity encountered by conventional contrastive losses, which heavily rely on margin and threshold value selection.

3.  A metric-induced decoder network is proposed to seamlessly integrate the refined bi-temporal features and generate the final change prediction under the guidance of the temporal relationship captured by metric distance modeling. The proposed network improves change feature modeling by leveraging the improved bi-temporal features learned by the hard sample-aware contrastive loss.

## II. RELATED WORKS

### A. Traditional Change Detection

Automatic CD from remote sensing images has been a prominent and ongoing research area due to its crucial role in monitoring dynamic Earth processes and supporting diverse applications such as high-precision mapping, damaged building detection, and urban expansion [4], [17], [18]. An intuitive approach to CD involves directly comparing the classification results of the bi-temporal images[19]. These post-classification comparison methods, however, are hindered by the error accumulation problem, where an 80% classification accuracy model results in only 64% accuracy after post-comparison. Consequently, the focus of CD research lies in jointly modeling temporal correlation and detecting change regions from multiple observations[20]–[22].

Early methods utilized algebraic approaches, such as differentiation, ratio, or vector analysis, to measure the spectral difference across different observations[23]. Due to the lack of spatial information in raw image pixels, the algebra-based methods are susceptible to inaccuracies caused by illumination variation, leading to pseudo-change detection. To address this issue, subsequent enhancements involved transforming bi-temporal images into features such as principal components and remote sensing indices before comparing their differences[24]. Considering the multiscale characteristics of remote sensing images, some studies have concentrated on the aggregation of multiscale features to enhance feature discriminability[10]. For example, Marchesi et al. utilized spectral change vector analysis at multiple scales to improve the model's robustness against registration noise[25]. Apart from developing improved feature descriptors, how to compare the feature difference has also attracted significant attention. For instance, Prendes et al. designed a statistical model to measure the similarity between bi-temporal remote sensing images by considering the sensors' physical properties[26]. Wu et al. introduced slow feature analysis to suppress unchanged pixels and achieve better separation of changed pixels. In [27], Markov random field is applied to model temporal relation from bi-temporal image pairs. As these early CD methods were primarily designed for medium and low-resolution remote sensing images, their ability is limited in generalizing to very high-resolution images. VHR images often exhibit richer high-frequency information and greater intra-object variability, which pose challenges to the applicability of these early methods.

*B. Segmentation-based DLCD*

The past decades have witnessed the flourishing of deep learning techniques and their application to remote sensing image interpretation. The deep learning-based CD(DLCD) has surpassed traditional CD methods in terms of performance, attributed to its superior representation and fitting capabilities. Early DLCD methods adopt early fusion strategies, wherein bi-temporal images were differenced or concatenated before being fed into the network. However, these strategies failed to effectively accommodate the key characteristics of change detection, making the models susceptible to pseudo-change detection. Further CD methods employed the Siamese architecture, in which a weight-sharing feature extractor is applied to process bi-temporal images separately. The Siamese network operates under the assumption that an object in homogeneous images should possess the same embedding. Consequently, changed regions can be detected based on the variations in their representations. It should be noted that any advanced feature extractors can be converted to the Siamese architecture for hierarchical feature extraction. Extensive studies have focused on how to convert the bi-temporal features into a change map. Inspired by semantic segmentation, the mainstream segmentation-based approaches introduce a decoder network to fuse the bi-temporal features and generate a change prediction based on multi-scale change features [9]. The change features are originally generated by concatenation[28] or absolute difference[29]. Subsequent research has focused on enhancing the presentation of change features during the fusion and decoder stages[2]. For example, Zhang et al. introduced a convolutional block attention module to fuse bi-temporal features[12]. Li et al. proposed A2Net and designed a supervised attention mechanism for better change feature representation[30]. Zheng et al.[31] designed a temporal-symmetric transformer to ensure symmetry of bi-temporal features and thus improve change feature representation. The transformer architecture has also been introduced in CD to capture global context information from images[7], [32]. However, segmentation-based methods encounter issues of insufficient temporal modeling as these methods are optimized using classification loss in the output space between the change map and the ground truth. While some studies introduced recurrent neural network architectures to process temporal information, the temporal relationship is implicitly modeled, and the computational costs associated with these approaches are considerably high. As a result, the alignment of object representations in the feature space cannot be guaranteed, potentially compromising change feature representation in segmentation-based methods. Therefore, we argue that further improvements can be achieved by explicit and meticulously modeling temporal features in a memory-efficient manner.

*C. Metric-based DLCD*

As modeling temporal correlation is a vital part of change detection, the metric-based methods model the metric distance of bi-temporal features in the metric space. Bi-temporal features extracted from the Siamese feature extractor are projected into the metric space; representations of the unchanged regions are pulled closer and those of the changed regions are pushed apart. Zhan et al. first introduced metric learning to the deep learning-based CD[14], in which a weighted contrastive loss is adopted to maximize feature distance in the changed regions and minimize distance in the non-changed regions. The threshold value with the best performance on the training sample is selected to generate the binary change map. To enhance interclass separability and intraclass inseparability in the feature space, a triplet loss has been proposed for CD[15]. Another approach, described in [33], aligns pyramid features of homogeneous regions in a consistent space, followed by the generation of a difference map through metric distance calculation between the output features of autoencoders[34].

Despite the strong temporal modeling capacity exhibited by metric-based DLCD methods, they are susceptible to suffering from pseudo-change detection and threshold selection problems. Since the binary change map is derived from the metric distance of bi-temporal features using manually chosen thresholds, the selection process becomes subjective, leading to the potential misclassification of misregistered regions as change categories. To mitigate the issue of pseudo-change detection, Chen et al.

TABLE I PSEUDOCODE OF HARD SAMPLE-AWARE CONTRASTIVE LOSS IN THE PYTORCH STYLE

**Algorithm1 Pseudocode of Hard Sample-aware Contrastive Loss**

#Siamese encoder network $N$ with weight $\theta$, input image $I_{T1}$ and $I_{T2}$, and temperature t

1. Extract features from bi-temporal images: $f_{T1}=N(I_1,\theta), f_{T2}=N(I_2,\theta),$

2. loss=BCEwithLogitsLoss(-[L2normalize($f_{T1}$)·L2normalize($f_{T1}$)].sum(dim=1)/t, labels)

3. Update the parameters of Siamese encoder network $N$ :loss.backward()

introduced self-attention mechanisms to model the spatial-temporal relations between bi-temporal features. Additionally, a contrastive loss is applied to optimize the Euclidean distance of the changed and non-changing features in the metric space[35]. Shi et al. further proposed the incorporation of auxiliary classifiers to provide deep supervision during the metric learning process[16]. Despite the efforts made to improve the metric-based DLCD methods, the CD paradigm of metric introduces the risk of introducing pseudo and unexpected changes in the final CD results. Therefore, it holds significance to complement the strong fitting abilities of segmentation-based DLCD with the robust temporal modeling of metric-based DLCD to improve overall CD performance.

## III. HARD SAMPLE-AWARE METRIC OPTIMIZATION

A hard sample-aware metric loss is designed to address the insufficient temporal modeling problem of the existing CD methods. We start with the introduction of contrastive loss[36], a fundamental objective function in metric-based CD methods. The motivation for contrastive loss lies in the assumption that features extracted from the same location for unchanged objects should exhibit similar representations, whereas features of changed objects should possess dissimilar representations. Given the bi-temporal features $f_{t1}$ and $f_{t2}$ extracted from the Siamese encoder network, the contrastive loss maximizes the feature distance of the changed pixels and minimizes the distance of the unchanged pixels, as follows:

$$L_{con}=\frac{1}{\sum(1-y)}\frac{1}{2}\sum_{c,h,w}(1-y)\cdot D(f_{t1},f_{t2})+$$
$$\frac{1}{\sum y}\frac{1}{2}\sum_{c,h,w}y\cdot Max\big(0,m-D(f_{t1},f_{t2})\big) \quad (1)$$

where y represents the ground truth label of 0 being unchanged and 1 being the change categories; $D(\cdot)$ denotes the distance metric such as Euclidean distance or Cosine distance. In $L_{con}$, the distance of changed features is enlarged until it is above the margin $m$, while the distance of the unchanged regions is optimized toward zero in the metric space. Afterward, pixels of feature distance above the threshold $d_{thre}$ are regarded as the changed category.

$$\hat{y}=\begin{cases}1, D>d_{thre}\\0, D\leq d_{thre}\end{cases} \quad (2)$$

In Eq.1 and 2, the limitation of the contrastive loss is that it relies heavily on the selection of margin value $m$ and the distance threshold $d_{thre}$ , which are typically determined through heuristics or manual selection. Furthermore, the effectiveness of temporal relation modeling is compromised when learning from a large number of accurately classified non-changing samples and only a few hard samples, given that non-changed regions often occupy a significant portion of the image. In light of this, our proposed hard sample-aware contrastive loss aims to enhance the performance of temporal modeling by automatically decreasing the weight of gradients assigned to well-classified samples and directing the model's attention toward challenging examples. As depicted in Fig.2, the bi-temporal features are first projected onto the L2 hypersphere, and the cosine similarly distance can be calculated between the projected features as follows:

$$\begin{cases}u_{t1}=\frac{f_{t1}}{\|f_{t1}\|_2}\\u_{t2}=\frac{f_{t2}}{\|f_{t2}\|_2}\end{cases} \quad (3)$$

$$D_{cos}=1-\frac{1}{2}(u_{t1}-u_{t2})^2=\cos(f_{t1},f_{t2}) \quad (4)$$

A sigmoid operation $\sigma(\cdot)$ is applied to the temperature-scaled cosine similarity $D_{cos}$ , and a binary cross entropy loss is introduced to optimize temporal feature correlation:

$$L_{HSAC}=-\frac{1}{N}\sum_N(1-y)\log[1-\sigma\left(-\frac{D_{cos}}{\tau}\right)]+y\log\sigma\left(-\frac{D_{cos}}{\tau}\right)$$

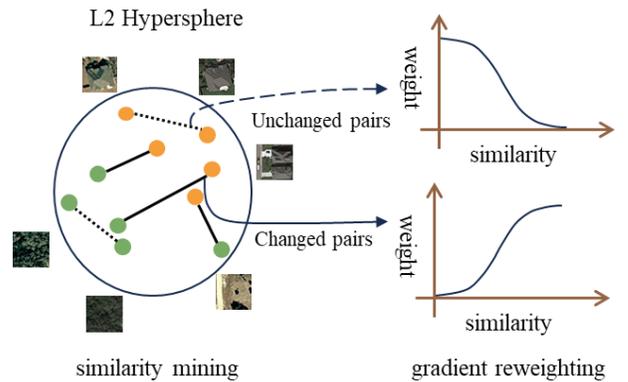

Fig.2 Illustration of the gradients weights under hard sample-aware contrastive loss.

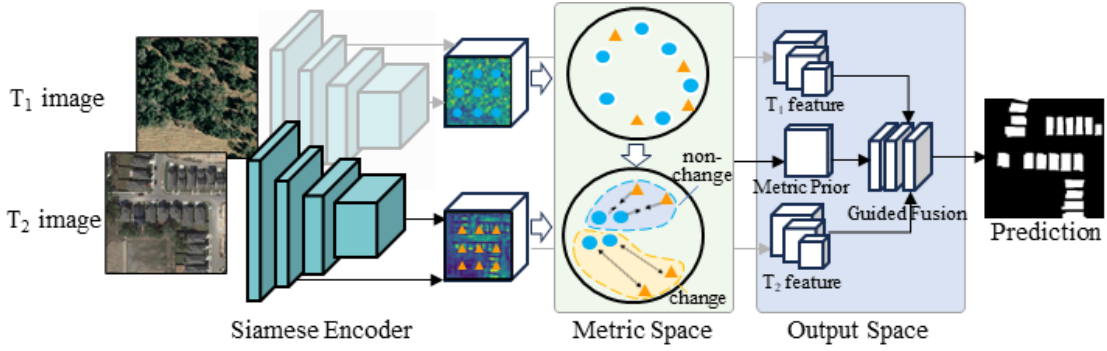

Fig.3 Overall Architecture of proposed deep change feature learning (DeepCL) framework.

$$= -\frac{1}{N}\Sigma_N (1-y)\log\sigma\left(\frac{D_{cos}}{\tau}\right) + y\log\left(1-\sigma\left(\frac{D_{cos}}{\tau}\right)\right) \quad (5)$$

In the unchanged regions, the objective function is $-\log\sigma\left(\frac{D_{cos}}{\tau}\right)$, where the loss is increased when $D_{cos}$ is smaller. Conversely, in the changed regions, the objective function follows $\log\left(1-\sigma\left(\frac{D_{cos}}{\tau}\right)\right)$, with the loss increasing as the cosine similarity rises. It should be noted that the implementation of $L_{HSAC}$ is straightforward with merely one line of code in implantation, as shown in PyTorch style pseudocode in Table 1.

We then explain why $L_{HSAC}$ is aware of the hard samples in the gradient aspects. According to the chain rule, the gradients of $L_{HSAC}$ is formulated as:

$$\frac{dL_{HSAC}}{d\theta} = \frac{dL_{HSAC}}{d\sigma} \cdot \frac{d\sigma}{dcos} \cdot \frac{dcos}{d\theta} \quad (6)$$

The components of which can be formulated as:

$$\frac{dL_{HSAC}}{d\sigma} = -\frac{1}{N}\Sigma_N \frac{1-y}{\sigma\left(\frac{D_{cos}}{\tau}\right)} - \frac{y}{1-\sigma\left(\frac{D_{cos}}{\tau}\right)} = -\frac{1}{N}\Sigma_N \frac{1-y-\sigma\left(\frac{D_{cos}}{\tau}\right)}{\sigma\left(\frac{D_{cos}}{\tau}\right)\cdot\left(1-\sigma\left(\frac{D_{cos}}{\tau}\right)\right)} \quad (7)$$

$$\frac{d\sigma}{dcos} = \frac{d(1+e^{-cos})^{-1}}{dcos} = \sigma\left(\frac{D_{cos}}{\tau}\right)\cdot\left[1-\sigma\left(\frac{D_{cos}}{\tau}\right)\right] \quad (8)$$

And thus

$$\frac{dL_{HSAC}}{d\theta} = -\frac{1}{N}\Sigma_N\left[1-y-\sigma\left(\frac{D_{cos}}{\tau}\right)\right]\cdot\frac{dcos}{d\theta} \quad (9)$$

in which

$$= \frac{dcos}{d\theta} = \frac{dcos}{df_{t1}}\cdot\frac{df_{t1}}{d\theta} + \frac{dcos}{df_{t2}}\cdot\frac{df_{t1}}{d\theta}$$
$$= \frac{f_{t1}}{\|f_{t1}\|_2\cdot\|f_{t2}\|_2}\frac{df_{t1}}{d\theta} + \frac{f_{t2}}{\|f_{t1}\|_2\cdot\|f_{t2}\|_2}\frac{df_{t2}}{d\theta} \quad (10)$$

We then discuss the gradients of $L_{HSAC}$ on the parameter $\theta$ of the Siamese encoder network:

$$\frac{dL_{HSAC}}{d\theta} = \begin{cases} \sigma\left(\frac{cos(f_{t1},f_{t2})}{\tau}\right)\cdot\frac{dcos}{d\theta}, y=1 \\ \left(1-\sigma\left(\frac{cos(f_{t1},f_{t2})}{\tau}\right)\right)\cdot\frac{dcos}{d\theta}, y=0 \end{cases} \quad (11)$$

The conventional contrastive loss simply formulates $\frac{dcos}{d\theta}$ as the gradients of the network parameters, resulting in an overwhelming dominance of well-classified easy samples. Consequently, this leads to inadequate model learning. In contrast, our proposed hard sample-aware contrastive loss effectively addresses this issue by automatically balancing the weights assigned to easy and hard samples. Specifically, in the unchanged regions where $y=0$, gradients of easy samples are down-weighted to 0 as $\left(1-\sigma\left(\frac{cos(f_{t1},f_{t2})}{\tau}\right)\right) \to 0$ due to the tendency of $cos(f_{t1},f_{t2}) \to 1$. In the false-negative regions where $y=1$ and $cos(f_{t1},f_{t2}) \to 1$, the weights of gradients are upscaled to 1, thereby enabling the model to focus on learning from the hard examples. The proposed hard sample-aware contrastive loss diminishes the impact of easy samples on the overall loss. To illustrate, consider a hard change sample with $cos(f_{t1},f_{t2})=0.5$, which would have gradient weights approximately 10 times higher than those assigned to an easy sample with $cos(f_{t1},f_{t2})=-0.5$. In contrast, the conventional contrastive loss assigns identical gradient weights to all samples, regardless of their level of difficulty. This uniform weighting scheme disregards the challenges posed by the abundance of easy non-changing samples, thereby leading to ineffective learning.

## IV. DeepCL Framework

DeepCL is a unified CD framework consisting of a Siamese encoder, a metic-space optimization mechanism, and an output space optimization mechanism. he Siamese encoder, as depicted in Fig. 3, extracts features from bi-temporal images The extracted features are then projected into a latent metric space, where the hard sample-aware contrastive loss function model temporal relationships. The cosine distance of the changed objects is pulled apart and the unchanged objects are pushed closer in the metric space. The gradients originating from hard, misclassified samples are up-weighted, directing the model's learning toward challenging examples. Furthermore, the modeled temporal relations serve as change prior information to guide the segmentation of change regions. The CD segmentation results are optimized using cross-entropy and dice loss in the output space. DeepCL harnesses the temporal modeling capability of metric-space optimization along with the strong fitting capability of output space optimization. On one hand, optimizing the model in the metric space enhances the feature representation and aids the decoder network in accurately delineating change regions. On the other hand, the decoder network automatically fits the CD ground truth, thereby mitigating issues related to pseudo-changes and manual threshold selection inherent in purely metric-based CD methods. Consequently, DeepCL emerges as a robust framework exhibiting superior feature representation, robustness against pseudo-changes, and applicability across diverse CD scenarios.

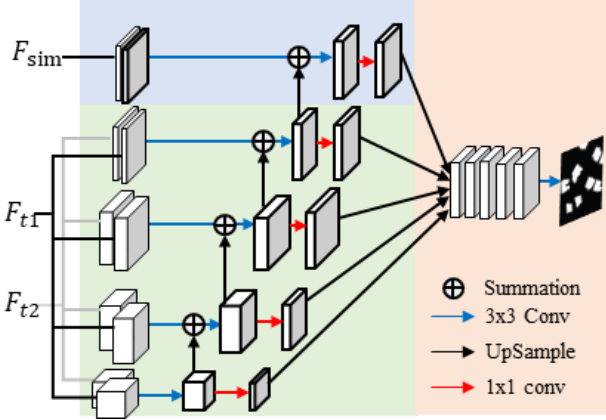

Fig.4 Metric-induced decoder module.

**Siamese Encoder :** The encoder part is responsible for extracting features from the input images. In the Siamese encoder network, identical network architectures with shared weights are applied to the bi-temporal features. A well-designed encoder is essential for capturing representative features, thus enhancing the performance of subsequent tasks. State-of-the-art encoder networks, such as VGGNet[37], ResNets[38], and EfficientNet[39] have demonstrated their effectiveness across various vision tasks. To leverage the merits of these superior network architectures, we have opted for EfficientNet as the encoder for bi-temporal feature extraction. The architecture of EfficientNet was designed by neural architecture search and thus can achieve a better tradeoff between speed and accuracy. It has five down-sampling stages with different spatial resolutions as feature output. We collect the extracted hierarchical features of the bi-temporal image for further CD. Given remote sensing images $I_t$ of phase $t$ the input, the output features $X_t$ can be obtained as:

$$F_t = \{f_i^i\}_{i=1}^5 \tag{12}$$

$f_i$ denotes the extracted feature with stride=2, thus the spatial resolution of $f_i$ is twice higher than $f_{i+1}$. To ensure computational efficiency, we have chosen EfficientNet-b0, which represents the most lightweight variant, as the encoder network.

**Metric Optimization:** To capture the temporal relationship, the extracted bi-features are up-sampled to the highest feature resolution. and are concatenated and projected into the cosine metric space as follows:

$$f_t = \left\| Conv_{3x3}([Up(f_i^i)]_{i=1}^5) \right\|_2 \tag{13}$$

The temporal correlation between the projected bi-temporal feature is optimized using the hard sample-aware contrastive loss described in section III, thus representations of the non-changed pixel pair are pushed closer while the changed pair is pulled apart in the metric space.

**Metric-induced Segmentation:** The existing metric-based CD methods employ a straightforward binary change mapping approach based on thresholding the metric distance between bi-temporal features. However, the resulting change map is prone to false detections of pseudo-change regions caused by factors such as image misregistration and shadow sheltering. Despite exhibiting visual dissimilarities in the images, these pseudo-change regions should not be identified as genuine change regions since they are false-alarmed. To address this limitation and enable the model to fully exploit the ground truth labels, we propose a metric-induced decoder module to segment change regions from the temporal relation-enhanced features. This simple architecture, as illustrated in Fig.4, does compare favorably to the more complex design. Instead of using only the bi-temporal features extracted by the encoder, we incorporate the projected metric features $f_{t1}$ and $f_{t2}$ to guide the model in accurately segmenting change regions amidst complex backgrounds. The hierarchical bi-temporal features are concatenated and fused using convolutional layers with a kernel size of 3x3. Afterward, these features are subjected to a cross-scale feature interaction process, which involves upsampling and integrating features from lower-resolution branches into higher-resolution branches. A light convolutional decoder head of kernel size (1,1) is applied to output features at different spatial resolutions respectively. These multi-scale features are subsequently used as input to a classification head and generate pixel-wise segmentation results.

## 2.1 Experimental Setup

**Dataset.** We validate the performance of DeepCL using three publicly available CD benchmarks, starting with an introduction of these benchmark datasets.

The LEVIR-CD dataset is a large-scale change detection dataset containing 637 image patches of size 1024x1024 pixels with the corresponding change labels. The dataset covers 20 distinct cities in the United States and is suitable for simulating the practical CD mapping scenario nationwide. The bi-temporal images captured span a period of 5-14 years, encompassing various types of building changes, including both new constructions and demolitions. The season and illumination variation between bi-temporal remote sensing images poses a great challenge to the models' performance on detecting real building change from pseudo changes. Since the original image size is too large for network training, we followed the common practice of existing studies and clipped the images and labels into 256x256 pixel patches without overlapping. As a result, we obtained a total of 7120, 1024, and 2048 patches for the training, validation, and test sets, respectively.

The WHU building change dataset focuses on the reconstruction of buildings after an earthquake occurred in Christchurch, New Zealand. The VHR remote sensing images are 0.3m in spatial resolution. The corresponding change labels include approximately 4,000 instances of changed buildings, making it a challenging task to accurately detect building changes across different scenes. In accordance with the official dataset partition, we obtained 4838 patches for training purposes and 2596 patches for testing.

The Google Earth dataset is a large-scale change detection dataset containing very high-resolution RS images of six cities situated in China. The remote sensing images are obtained from

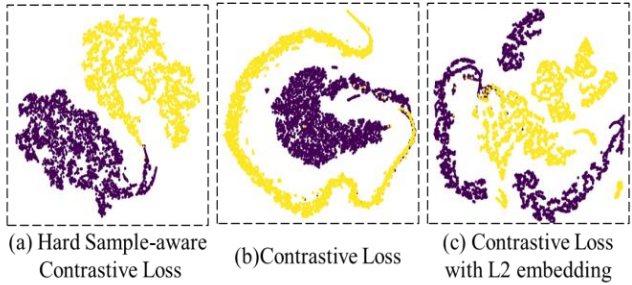

(a) Hard Sample-aware Contrastive Loss    (b)Contrastive Loss    (c) Contrastive Loss with L2 embedding

Fig.5 T-SNE visualization of embeddings of bi-temporal change pixel pair.

Google Earth with a ground sampling distance of 0.3m. The training dataset is composed of 3600 image patches that cover five cities across China; the test dataset is composed of 480 image patches exclusively obtained from a distinct city. As the training and test areas originate from different regions, the Google Earth dataset is suitable for evaluating the model's generalization abilities in the context of large-scale CD mapping.

**Evaluation Metrics.** We adopted four evaluation metrics, including the intersection over union (IoU), precision, recall, and F1-score to evaluate the performance of the models. Given the predicted CD maps and its corresponding ground truth, we calculate the confusion metric including the number of true-positive(TP), true-negative(TN), false-positive(FP), and false-negative(FN) pixels, and calculate the evaluation metrics as follows:

$$IoU = \frac{TP}{TP+FN+FP} \tag{14}$$

$$Precision = \frac{TP}{TP+FP} \tag{15}$$

$$Recall = \frac{TP}{TP+FN} \tag{16}$$

$$F1 = \frac{2 \times precision \times recall}{precision+recall} \tag{17}$$

**Optimization:** DeepCL was trained with the AdamW optimizer. We used an RTX 3090 GPU with a total of 12 images per minibatch. The model was trained for 100 epochs with an initial learning rate of 0.001, which was then reduced using a polynomial learning rate policy of power 0.9. A weight decay of 0.01 was employed to regulate the model's weights. To enhance data diversity and improve model robustness, online data augmentation techniques were employed, including random rotation and flipping. The optimization process of DeepCL consisted of both metric-space optimization and output-space optimization. The metric-space optimization focuses on modeling temporal relationships using the proposed hard sample-aware contrastive loss; the output space optimization comprises a combination of Dice loss and cross-entropy loss between the segmentation outcome and the ground truth label.

## V. EXPERIMENTS

We conduct experiments to analyze the behavior of the proposed hard sample-aware metric optimization and various metric loss functions for metric-based CD. Afterward, we analyze how metric-induced segmentation further improves the existing metric-based CD performance. Finally, we compare the performance of the components in the DeepCL framework with the state-of-the-art CNN-based and Transformer-based CD architectures.

TABLE II PERFORMANCE OF METRIC-BASED CD METHODS UNDER DIFFERENT LOSS FUNCTIONS

| Dataset | Loss | IoU | F1-score |
|---------|------|-----|----------|
| LEVIR-CD | $L_{con}$ | 76.96 | 86.98 |
| | $L_{conL2}$ | 75.58 | 86.09 |
| | $L_{HSAC}$ | **82.44** | **90.37** |
| WHU-CD | $L_{con}$ | 82.41 | 90.36 |
| | $L_{conL2}$ | 81.28 | 89.67 |
| | $L_{HSAC}$ | **84.99** | **91.89** |
| Google | $L_{con}$ | 47.16 | 64.09 |
| | $L_{conL2}$ | 49.15 | 65.91 |
| | $L_{HSAC}$ | **50.69** | **67.28** |

### A. Performance of metric-space optimization

To evaluate the effectiveness of the proposed hard sample-aware contrastive loss, we compare its performance with two existing contrastive loss functions: contrastive loss calculated from the original feature space ($L_{con}$), and contrastive loss with features projected in the L2 metric space ( $L_{conL2}$ ). The experimental results presented in Table II demonstrate that the proposed metric loss outperforms the original contrastive loss across all three datasets. $L_{HSAC}$ exceeds $L_{con}$ by 5.48%, 2.58%, and 3.53 IoU on the LEVIR-CD, WHU-CD, and Google-CD datasets, respectively. Furthermore, we visualize the distribution of bi-temporal features in the changed pair using the T-SNE technique, as depicted in Fig.5. The visualization illustrates that metric space optimization with $L_{HSAC}$ achieves the most distinct separation of change pairs among bi-temporal features, showcasing the improved manifold structure and feature representation learned by $L_{HSAC}$. $L_{HSAC}$ successfully mitigates the issue of the overwhelming dominance of well-classified easy samples in model gradient, thereby directing its focus towards more challenging samples. Consequently, it effectively facilitates model learning within the metric space.

Next, we investigate the applicability of hard sample-aware contrastive loss to the existing metric-based CD methods. Specifically, we evaluate the performance of the proposed loss function in conjunction with three existing metric-based CD methods, namely, DASNet[40], STANet[35], and DSAMNet[16]. As shown in Table III, yields notable improvements in the performance of these methods, with a minimum IoU increment of 2.57% and an F1-score increment of 1.28%. It demonstrates the generalization ability of $L_{HSAC}$ in effectively modeling temporal relationships between bi-temporal images.

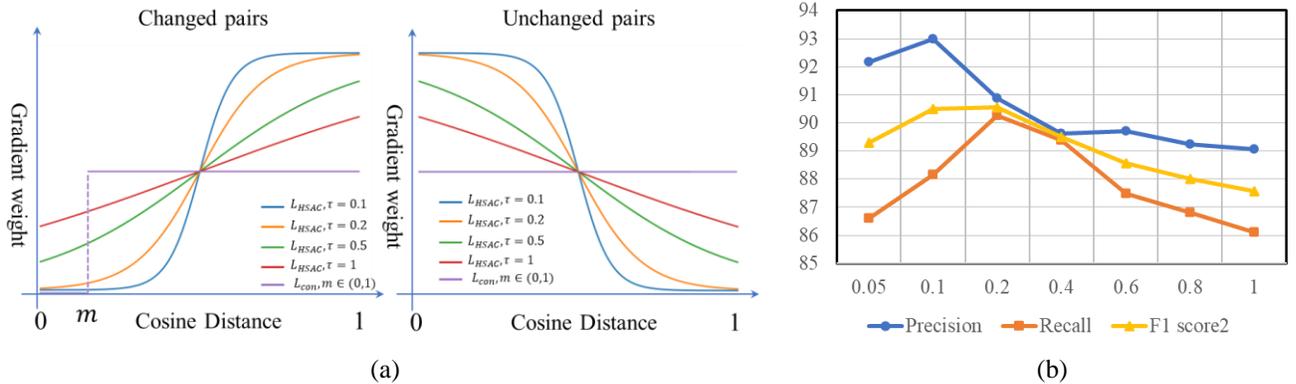

Fig. 6 (a) Gradient weights of hard sample-aware contrastive loss under different temperature values $\tau$. (b) Model performance under different temperature values $\tau$.

We conduct experiments of $L_{HSAC}$ under different temperature values, as shown in Fig.6 (b). We can see that the model's precision value gradually drops with the temperature increases. The recall value peaks when the temperature is equal to 0.2, in which the model can achieve the best F1-score with the best balance between precision and recall value. Thus, we adopt $\tau = 0.2$ as the default setting.

The selection of temperature significantly impacts the gradient weights and the overall performance of the model. Fig. 6(a) provides a visualization of the gradient weights in $L_{HSAC}$ under various temperature values. It is observed that when the temperature approaches zero, hard change samples with a cosine similarity close to 1 and hard unchanged samples with a cosine similarity close to 0 are heavily penalized, while the weights assigned to easy samples are suppressed. As the temperature value increases, the degree of penalization is gradually reduced, allowing the model to also learn from semi-hard samples. As the temperature value tends towards infinity, $L_{HSAC}$ devolves into the original contrastive loss, where both hard and easy samples share the same gradient weight.

To investigate the impact of temperature selection, we conducted experiments with $L_{HSAC}$ using different temperature values, as depicted in Fig. 6(b). Notably, as the temperature increases, the model's precision value experiences a gradual decline. However, the recall value reaches its peak when the temperature is set to 0.2, which enables the model to attain the highest F1 score by the optimal balance between precision and recall. Therefore, a value of $\tau$=0.2 is adopted as the default setting for the temperature hyperparameter in $L_{HSAC}$.

### B. Ablation Experiments

To validate the components of DeepCL, ablation experiments were conducted, with the experimental settings outlined in Table IV. Given EfficientNet as the Siamese encoder backbone, we compare the performance gain of replacing the contrastive loss $L_{con}$ with hard sample-aware contrastive loss $L_{HSAC}$ in the metric-based CD manner. The experimental results displayed in Table V lines 1-2 demonstrate that the proposed hard sample-aware contrastive loss consistently outperforms the original contrastive loss across all datasets.

Furthermore, the performance of the proposed decoder in DeepCL was compared with the widely adopted U-Net-like decoder for segmentation-based CD methods. Ablation experiments 3 and 4 revealed that the proposed decoder surpassed the U-Net-like decoder due to its effective cross-scale fusion design. Additionally, the incorporation of metric-space optimization into the segmentation method further enhanced the model's performance in explicitly modeling temporal relations.

The DeepCL framework, incorporating the hard sample-aware contrastive loss and metric-induced segmentation, achieved superior performance across all three datasets, surpassing all other ablation settings.

### C. Comparison to State-of-the-art

To assess the effectiveness of our proposed DeepCL framework, we conduct a comprehensive comparison with state-of-the-art CD methods. Herein, we provide a concise description of these comparison methods, followed by quantitative and qualitative evaluations of the experimental outcomes.

**Comparison Methods.** We selected 11 state-of-the-art CD methods for comparison, including both advanced metric-based and segmentation-based methods. The metric-based CD methods, including DASNet[40], STANet[35], and DSAMNet[16] were selected for comparison. The segmentation-based methods includes FC-EF[9], FCSiam-Diff[9], FCSiam-Conc [9], MSPSNet[41], SNUNet[13], ChangeFormer[42], BIT[43], and PA-Former[44] for comparison. It should be noted that not only the CNN-based but also the recent transformer-based CD algorithms[42]–[44] were selected for comparison. We re-implemented these comparison methods using their publicly available codes under the recommended hyperparameter settings.

TABLE III PERFORMANCE OF METRIC-BASED CD METHODS UNDER DIFFERENT LOSS FUNCTIONS ON THE LEVIR-CD DATASET

| Method | Loss | IoU | F1-score |
|--------|------|-----|----------|
| DASNet | $L_{con}$ | 70.74 | 82.86 |
| | $L_{HSAC}$ | 73.84 (+3.1) | 84.95 (+2.09) |
| STANet | $L_{con}$ | 78.69 | 88.08 |
| | $L_{HSAC}$ | 81.26 (+2.57) | 89.66 (+1.58) |
| DSAMNet | $L_{con}$ | 74.11 | 85.13 |
| | $L_{HSAC}$ | 82.12 (+8.01) | 90.18 (+5.05) |

TABLE IV ABLATION EXPERIMENTAL SETTINGS OF DEEPCL

| Method | $L_{con}$ | $L_{HSAC}$ | D | MD | Description |
|---|---|---|---|---|---|
| Description | Contrastive Loss | Hard Sample-aware Loss | Simple decoder | Metric-induced decoder | |
| Ablation 1 | √ | | | | Metric-based CD with the original contrastive loss |
| Ablation 2 | | √ | | | Metric-based CD with hard Sample-aware loss |
| Ablation 3 | | | √ | | Segmentation-based CD with U-Net decoder |
| Ablation 4 | | | | √ | Segmentation-based CD with the proposed decoder |
| Ablation 5 | √ | | | √ | Combine contrastive loss with U-Net segmentation |
| Ablation 6 | | √ | | √ | Hard sample-aware loss with metric-induced Segmentation |

**Quantitative Evaluation.** The experimental results of DeepCL and the comparison methods are presented in Table VI, with the highest score denoted in bold and the second-highest score underlined. Notably, DeepCL demonstrates superior performance across all three datasets, surpassing all other comparison methods in terms of both the IoU and F1-score. In general, metric-based methods exhibit relatively lower performance compared to segmentation-based CD methods. This can be attributed to the inherent limitation of metric-based methods in terms of the lack of bi-temporal feature interaction, which may result in the misclassification of pseudo-change regions as actual change regions. Among the segmentation-based methods, the transformer-based CD methods show comparable performance with the CNN-based methods. By combining metric learning and segmentation approaches,

DeepCL achieves superior performance and surpasses all existing CD methods that focus on either metric-based or segmentation-based techniques.

**Visualization.** Furthermore, we visualize some examples of the model predictions of the three datasets in Fig.7, in which pixels in green, blue, and red denote the correctly classified, omitted, and misclassified change regions respectively. From Fig.7 column 3 we can see that DeepCL makes the most correct classification than the comparison methods. Within rows 1 and 5, it is apparent that many of the comparison methods struggle to accurately extract change regions in complex building structures. However, DeepCL successfully captures these changes by leveraging the combined advantages of metric learning and segmentation. In rows 3-4, it can be observed that many comparison methods, especially the metric-based methods, are prone to misclassify the pseudo changes regions caused by illumination difference and misregistration error as real change. However, DeepCL introduces the segmentation technique to fit the distribution of real change regions under the guidance of metric features, effectively suppressing the occurrence of pseudo-change regions. Furthermore, DeepCL exhibits a higher precision in predicting the edges of dense change regions compared to the comparison methods. This outcome demonstrates the efficacy of the cross-scale interaction design within the metric-induced segmentation module of DeepCL.

TABLE V ABLATION EXPERIMENTAL RESULTS OF DEEPCL

| Dataset<br>Method | LEVIR-CD | | WHU-CD | | Google | |
|---|---|---|---|---|---|---|
| | IoU(%) | F1-score(%) | IoU(%) | F1-score(%) | IoU(%) | F1-score(%) |
| Encoder+$L_{con}$ | 76.96 | 86.98 | 82.41 | 90.36 | 47.16 | 64.09 |
| Encoder+$L_{HSAC}$ | 82.44 | 90.37 | 84.99 | 91.89 | 50.69 | 67.28 |
| Encoder+D | 82.55 | 90.44 | 85.15 | 91.98 | 47.89 | 64.76 |
| Encoder+MD | 82.85 | 90.62 | 86.75 | 92.9 | 49.69 | 65.49 |
| Encoder+$L_{con}$+MD | 83.28 | 90.88 | 87.02 | 93.06 | 51.78 | 68.23 |
| Encoder+$L_{HSAC}$+MD | **83.68** | **91.11** | **87.4** | **93.28** | **53.31** | **69.55** |

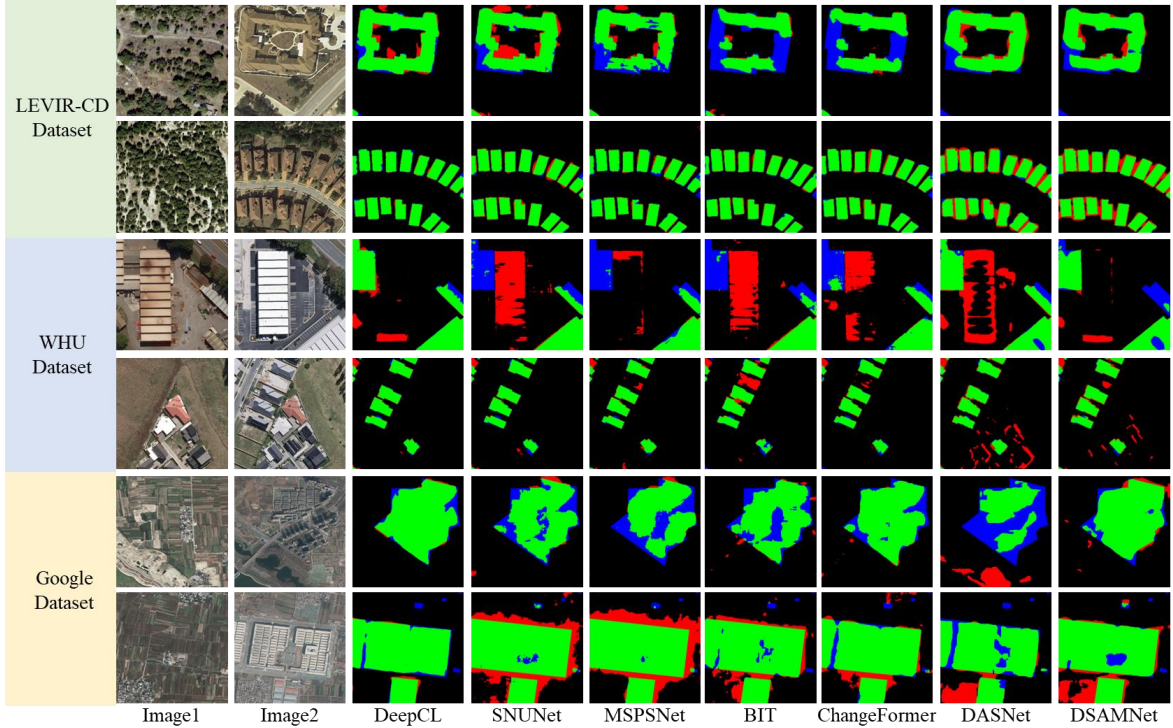

Fig.7. Visualization of the prediction results on the LEVIR-CD, WHU-CD, and Google datasets.

## VI. CONCLUSION

In this paper, a hard sample-aware contrastive loss is proposed to model temporal correlation from bi-temporal images with hard sample reweighted. Building upon the robust temporal modeling capabilities of optimization in the metric space, our proposed DeepCL architecture further leverages the strong fitting abilities of optimization in the output space to address the challenge of pseudo-change detection. Theoretical and experimental analyses substantiate the effectiveness of the hard sample-aware contrastive loss in enhancing feature discriminability compared to conventional contrastive loss techniques. Furthermore, the integration of metric learning and metric-induced segmentation within the DeepCL framework yields superior performance in comparison to state-of-the-art change detection methods, exhibiting lower rates of omission and pseudo-change misclassification. Consequently, DeepCL is an effective framework that combines the strong fitting capabilities of segmentation and the robust temporal modeling of metric learning, contributing to explainable and robust change detection. Future research endeavors could investigate explicit modeling techniques accounted for temporal correlations in multi-temporal remote sensing images, thereby facilitating long time-series Earth surface monitoring.

TABLE VI QUANTITATIVE EXPERIMENTAL RESULTS OF THE COMPARISON METHODS

| Method | Type | LEVIR-CD Dataset | | | | WHU-CD Dataset | | | | Google Dataset | | | |
|---|---|---|---|---|---|---|---|---|---|---|---|---|---|
| | | IoU | Precision | Recall | F1-score | IoU | Precision | Recall | F1-score | IoU | Precision | Recall | F1-score |
| DASNet | | 70.74 | 77.3 | 89.28 | 82.86 | 75.1 | 88.72 | 83.02 | 85.78 | 43.0 | 60.41 | 59.89 | 60.14 |
| DSAMNet | Metric | 74.11 | 82.26 | 88.21 | 85.13 | 75.15 | 84.48 | 87.19 | 85.81 | 51.36 | 61.48 | 75.72 | 67.86 |
| STANet | | 74.54 | 78.94 | **93.05** | 85.42 | 72.99 | 81.47 | 87.52 | 84.38 | 32.29 | **80.32** | 35.07 | 48.82 |
| FC-EF | | 74.9 | 86.9 | 84.43 | 85.65 | 76.98 | 89.43 | 84.69 | 86.99 | 33.33 | 50.31 | 49.69 | 50.0 |
| FCSiam-Diff | | 80.19 | 91.31 | 86.81 | 89.01 | 77.88 | 89.17 | 86.02 | 87.57 | 41.48 | 59.55 | 57.76 | 58.64 |
| FCSiam-Conc | | 70.7 | 87.26 | 78.84 | 82.84 | 71.51 | 85.47 | 81.4 | 83.39 | 46.01 | 63.51 | 62.55 | 63.03 |
| MSPSNet | Segmen- | 83.0 | **91.92** | 89.53 | 90.71 | 82.32 | 92.39 | 88.31 | 90.3 | 50.56 | 63.03 | 71.87 | 67.16 |
| SNUNet | tation | 82.55 | 90.82 | 90.07 | 90.44 | 81.24 | 91.12 | 88.22 | 89.65 | 53.19 | 62.29 | **78.45** | 69.44 |
| ChangeFormer | | 82.29 | 92.29 | 88.37 | 90.29 | 78.44 | 92.77 | 83.54 | 87.92 | 50.2 | 59.02 | 77.06 | 66.84 |
| PAFormer | | 82.31 | 91.82 | 88.82 | 90.29 | 83.47 | 91.52 | 90.47 | 90.99 | 49.12 | 62.04 | 70.22 | 65.88 |
| BIT | | 81.94 | 91.77 | 88.44 | 90.07 | 70.77 | 77.96 | 88.47 | 82.88 | 50.07 | 72.97 | 61.48 | 66.73 |
| DeepCL | Combine | **83.68** | 91.84 | 90.39 | **91.11** | **87.4** | **95.77** | 90.9 | **93.28** | **53.31** | 69.56 | 69.54 | **69.55** |

## REFERENCES

[1] X. X. Zhu *et al.*, 'Deep Learning in Remote Sensing: A Comprehensive Review and List of Resources', *IEEE Geoscience and Remote Sensing Magazine*, vol. 5, no. 4, pp. 8–36, Dec. 2017, doi: 10.1109/MGRS.2017.2762307.

[2] T. Bai *et al.*, 'Deep learning for change detection in remote sensing: a review', *Geo-spatial Information Science*, vol. 0, no. 0, pp. 1–27, Jul. 2022, doi: 10.1080/10095020.2022.2085633.

[3] M. Lin, G. Yang, and H. Zhang, 'Transition Is a Process: Pair-to-Video Change Detection Networks for Very High Resolution Remote Sensing Images', *IEEE Transactions on Image Processing*, vol. 32, pp. 57–71, 2023, doi: 10.1109/TIP.2022.3226418.

[4] Z. Zheng, Y. Zhong, J. Wang, A. Ma, and L. Zhang, 'Building damage assessment for rapid disaster response with a deep object-based semantic change detection framework: From natural disasters to man-made disasters', *Remote Sensing of Environment*, vol. 265, p. 112636, Nov. 2021, doi: 10.1016/j.rse.2021.112636.

[5] H. Guo, Q. Shi, A. Marinoni, B. Du, and L. Zhang, 'Deep building footprint update network: A semi-supervised method for updating existing building footprint from bi-temporal remote sensing images', *Remote Sensing of Environment*, vol. 264, p. 112589, Oct. 2021, doi: 10.1016/j.rse.2021.112589.

[6] M. Decuyper *et al.*, 'Continuous monitoring of forest change dynamics with satellite time series', *Remote Sensing of Environment*, vol. 269, p. 112829, Feb. 2022, doi: 10.1016/j.rse.2021.112829.

[7] M. Liu, Z. Chai, H. Deng, and R. Liu, 'A CNN-Transformer Network With Multiscale Context Aggregation for Fine-Grained Cropland Change Detection', *IEEE Journal of Selected Topics in Applied Earth Observations and Remote Sensing*, vol. 15, pp. 4297–4306, 2022, doi: 10.1109/JSTARS.2022.3177235.

[8] M. Hu, C. Wu, B. Du, and L. Zhang, 'Binary Change Guided Hyperspectral Multiclass Change Detection', *IEEE Transactions on Image Processing*, vol. 32, pp. 791–806, 2023, doi: 10.1109/TIP.2022.3233187.

[9] R. Caye Daudt, B. Le Saux, and A. Boulch, 'Fully Convolutional Siamese Networks for Change Detection', in *2018 25th IEEE International Conference on Image Processing (ICIP)*, Oct. 2018, pp. 4063–4067. doi: 10.1109/ICIP.2018.8451652.

[10] H. Zhang, M. Lin, G. Yang, and L. Zhang, 'ESCNet: An End-to-End Superpixel-Enhanced Change Detection Network for Very-High-Resolution Remote Sensing Images', *IEEE Transactions on Neural Networks and Learning Systems*, pp. 1–15, 2021, doi: 10.1109/TNNLS.2021.3089332.

[11] S. Tian, Y. Zhong, Z. Zheng, A. Ma, X. Tan, and L. Zhang, 'Large-scale deep learning based binary and semantic change detection in ultra high resolution remote sensing imagery: From benchmark datasets to urban application', *ISPRS Journal of Photogrammetry and Remote Sensing*, vol. 193, pp. 164–186, Nov. 2022, doi: 10.1016/j.isprsjprs.2022.08.012.

[12] C. Zhang *et al.*, 'A deeply supervised image fusion network for change detection in high resolution bi-temporal remote sensing images', *ISPRS Journal of Photogrammetry and Remote Sensing*, vol. 166, pp. 183–200, Aug. 2020, doi: 10.1016/j.isprsjprs.2020.06.003.

[13] S. Fang, K. Li, J. Shao, and Z. Li, 'SNUNet-CD: A Densely Connected Siamese Network for Change Detection of VHR Images', *IEEE Geoscience and Remote Sensing Letters*, vol. 19, pp. 1–5, 2022, doi: 10.1109/LGRS.2021.3056416.

[14] Y. Zhan, K. Fu, M. Yan, X. Sun, H. Wang, and X. Qiu, 'Change Detection Based on Deep Siamese Convolutional Network for Optical Aerial Images', *IEEE Geoscience and Remote Sensing Letters*, vol. 14, no. 10, pp. 1845–1849, Oct. 2017, doi: 10.1109/LGRS.2017.2738149.

[15] M. Zhang, G. Xu, K. Chen, M. Yan, and X. Sun, 'Triplet-Based Semantic Relation Learning for Aerial Remote Sensing Image Change Detection', *IEEE Geoscience and Remote Sensing Letters*, vol. 16, no. 2, pp. 266–270, 2019, doi: 10.1109/LGRS.2018.2869608.

[16] Q. Shi, M. Liu, S. Li, X. Liu, F. Wang, and L. Zhang, 'A Deeply Supervised Attention Metric-Based Network and an Open Aerial Image Dataset for Remote Sensing Change Detection', *IEEE Transactions on Geoscience and Remote Sensing*, vol. 60, pp. 1–16, 2022, doi: 10.1109/TGRS.2021.3085870.

[17] Q. Zhu *et al.*, 'Land-Use/Land-Cover change detection based on a Siamese global learning framework for high spatial resolution remote sensing imagery', *ISPRS Journal of Photogrammetry and Remote Sensing*, vol. 184, pp. 63–78, Feb. 2022, doi: 10.1016/j.isprsjprs.2021.12.005.

[18] H. Guo, Q. Shi, A. Marinoni, B. Du, and L. Zhang, 'Deep building footprint update network: A semi-supervised method for updating existing building footprint from bi-temporal remote sensing images', *Remote Sensing of Environment*, vol. 264, p. 112589, Oct. 2021, doi: 10.1016/j.rse.2021.112589.

[19] Allan *et al.*, 'Multivariate Alteration Detection (MAD) and MAF Postprocessing in Multispectral, Bitemporal Image Data: New Approaches to Change Detection Studies', *Remote Sensing of Environment*, 1998, doi: 10.1016/S0034-4257(97)00162-4.

[20] M. Liu, Q. Shi, A. Marinoni, D. He, X. Liu, and L. Zhang, 'Super-Resolution-Based Change Detection Network With Stacked Attention Module for Images With Different Resolutions', *IEEE Transactions on Geoscience and Remote Sensing*, pp. 1–18, 2021, doi: 10.1109/TGRS.2021.3091758.

[21] D. Amitrano, R. Guida, and P. Iervolino, 'Semantic Unsupervised Change Detection of Natural Land Cover With Multitemporal Object-Based Analysis on SAR Images', *IEEE Transactions on Geoscience and Remote Sensing*, pp. 1–21, 2020, doi: 10.1109/TGRS.2020.3029841.

[22] Z. Zheng, A. Ma, L. Zhang, and Y. Zhong, 'Change is Everywhere: Single-Temporal Supervised Object Change Detection in Remote Sensing Imagery', *arXiv:2108.07002 [cs]*, Aug. 2021, Accessed: Apr. 12, 2022. [Online]. Available: http://arxiv.org/abs/2108.07002

[23] A. Alonso-González, C. López-Martínez, K. P. Papathanassiou, and I. Hajnsek, 'Polarimetric SAR Time Series Change Analysis Over Agricultural Areas', *IEEE Transactions on Geoscience and Remote Sensing*, vol. 58, no. 10, pp. 7317–7330, 2020, doi: 10.1109/TGRS.2020.2981929.

[24] G. F. Byrne, P. F. Crapper, and K. K. Mayo, 'Monitoring land cover changes by principal components analysis of multitemporal Landsat data. Remote. Envir. 10:175-184', *Remote Sensing of Environment*, vol. 10, no. 3, pp. 175–184, 1980, doi: 10.1016/0034-4257(80)90021-8.

[25] S. Marchesi, F. Bovolo, and L. Bruzzone, 'A Context-Sensitive Technique Robust to Registration Noise for Change Detection in VHR Multispectral Images', *IEEE Transactions on Image Processing*, vol. 19, no. 7, pp. 1877–1889, Jul. 2010, doi: 10.1109/TIP.2010.2045070.

[26] J. Prendes, M. Chabert, F. Pascal, A. Giros, and J.-Y. Tourneret, 'A New Multivariate Statistical Model for Change Detection in Images Acquired by Homogeneous and Heterogeneous Sensors', *IEEE Transactions on Image Processing*, vol. 24, no. 3, pp. 799–812, Mar. 2015, doi: 10.1109/TIP.2014.2387013.

[27] R. Touati, M. Mignotte, and M. Dahmane, 'Multimodal Change Detection in Remote Sensing Images Using an Unsupervised Pixel Pairwise-Based Markov Random Field Model', *IEEE Transactions on Image Processing*, vol. 29, pp. 757–767, 2020, doi: 10.1109/TIP.2019.2933747.

[28] C. Han, C. Wu, H. Guo, M. Hu, and H. Chen, 'HANet: A Hierarchical Attention Network for Change Detection With Bitemporal Very-High-Resolution Remote Sensing Images', *IEEE Journal of Selected Topics in Applied Earth Observations and Remote Sensing*, vol. 16, pp. 3867–3878, 2023, doi: 10.1109/JSTARS.2023.3264802.

[29] S. Zhu, Y. Song, Y. Zhang, and Y. Zhang, 'ECFNet: A Siamese Network With Fewer FPs and Fewer FNs for Change Detection of Remote-Sensing Images', *IEEE Geoscience and Remote Sensing Letters*, vol. 20, pp. 1–5, 2023, doi: 10.1109/LGRS.2023.3238553.

[30] Z. Li *et al.*, 'Lightweight Remote Sensing Change Detection with Progressive Feature Aggregation and Supervised Attention', *IEEE*

*Transactions on Geoscience and Remote Sensing*, pp. 1–1, 2023, doi: 10.1109/TGRS.2023.3241436.

[31] Z. Zheng, Y. Zhong, S. Tian, A. Ma, and L. Zhang, 'ChangeMask: Deep multi-task encoder-transformer-decoder architecture for semantic change detection', *ISPRS Journal of Photogrammetry and Remote Sensing*, vol. 183, pp. 228–239, Jan. 2022, doi: 10.1016/j.isprsjprs.2021.10.015.

[32] Q. Li, R. Zhong, X. Du, and Y. Du, 'TransUNetCD: A Hybrid Transformer Network for Change Detection in Optical Remote-Sensing Images', *IEEE Transactions on Geoscience and Remote Sensing*, vol. 60, pp. 1–19, 2022, doi: 10.1109/TGRS.2022.3169479.

[33] 'DPFL-Nets: Deep Pyramid Feature Learning Networks for Multiscale Change Detection | IEEE Journals & Magazine | IEEE Xplore'. https://ieeexplore.ieee.org/document/9439971 (accessed May 23, 2023).

[34] Y. Wu, J. Li, Y. Yuan, A. K. Qin, Q.-G. Miao, and M.-G. Gong, 'Commonality Autoencoder: Learning Common Features for Change Detection From Heterogeneous Images', *IEEE Transactions on Neural Networks and Learning Systems*, vol. 33, no. 9, pp. 4257–4270, Sep. 2022, doi: 10.1109/TNNLS.2021.3056238.

[35] H. Chen and Z. Shi, 'A Spatial-Temporal Attention-Based Method and a New Dataset for Remote Sensing Image Change Detection', *Remote Sensing*, vol. 12, no. 10, Art. no. 10, Jan. 2020, doi: 10.3390/rs12101662.

[36] R. Hadsell, S. Chopra, and Y. LeCun, 'Dimensionality Reduction by Learning an Invariant Mapping', in *2006 IEEE Computer Society Conference on Computer Vision and Pattern Recognition (CVPR'06)*, Jun. 2006, pp. 1735–1742. doi: 10.1109/CVPR.2006.100.

[37] K. Simonyan and A. Zisserman, 'Very Deep Convolutional Networks for Large-Scale Image Recognition', *arXiv:1409.1556 [cs]*, Apr. 2015, Accessed: Jun. 03, 2021. [Online]. Available: http://arxiv.org/abs/1409.1556

[38] K. He, X. Zhang, S. Ren, and J. Sun, 'Deep Residual Learning for Image Recognition', *arXiv:1512.03385 [cs]*, Dec. 2015, Accessed: Nov. 05, 2020. [Online]. Available: http://arxiv.org/abs/1512.03385

[39] M. Tan and Q. V. Le, 'EfficientNet: Rethinking Model Scaling for Convolutional Neural Networks', *arXiv:1905.11946 [cs, stat]*, Sep. 2020, Accessed: Jul. 13, 2021. [Online]. Available: http://arxiv.org/abs/1905.11946

[40] J. Chen *et al.*, 'DASNet: Dual Attentive Fully Convolutional Siamese Networks for Change Detection in High-Resolution Satellite Images', *IEEE Journal of Selected Topics in Applied Earth Observations and Remote Sensing*, vol. 14, pp. 1194–1206, 2021, doi: 10.1109/JSTARS.2020.3037893.

[41] Q. Guo, J. Zhang, S. Zhu, C. Zhong, and Y. Zhang, 'Deep Multiscale Siamese Network With Parallel Convolutional Structure and Self-Attention for Change Detection', *IEEE Transactions on Geoscience and Remote Sensing*, vol. 60, pp. 1–12, 2022, doi: 10.1109/TGRS.2021.3131993.

[42] W. G. C. Bandara and V. M. Patel, 'A Transformer-Based Siamese Network for Change Detection'. arXiv, Sep. 01, 2022. doi: 10.48550/arXiv.2201.01293.

[43] H. Chen, Z. Qi, and Z. Shi, 'Remote Sensing Image Change Detection With Transformers', *IEEE Transactions on Geoscience and Remote Sensing*, vol. 60, pp. 1–14, 2022, doi: 10.1109/TGRS.2021.3095166.

[44] M. Liu, Q. Shi, Z. Chai, and J. Li, 'PA-Former: Learning Prior-Aware Transformer for Remote Sensing Building Change Detection', *IEEE Geoscience and Remote Sensing Letters*, vol. 19, pp. 1–5, 2022, doi: 10.1109/LGRS.2022.3200396.